\pdfoutput=1

\documentclass[11pt]{article}

\usepackage[final]{acl}

\usepackage{times}
\usepackage{latexsym}
\usepackage{multirow}
\usepackage{booktabs}
\usepackage{tabularx}

\usepackage[export]{adjustbox}

\usepackage[T1]{fontenc}

\usepackage[utf8]{inputenc}

\usepackage{microtype}

\usepackage{inconsolata}
\usepackage{tikz}
\usepackage{tikz-qtree}
\usepackage{tikz}
\usetikzlibrary{matrix,decorations.pathreplacing,positioning,fit}
\usetikzlibrary{fit,positioning, decorations.pathreplacing}

\usepackage[inline]{enumitem}
\usepackage{adjustbox}
\usepackage{multirow}
\usepackage{newtxmath}
\usepackage{subcaption}
\renewcommand{\arraystretch}{0.8} 
\usepackage{graphicx}
\usepackage{xcolor}

\definecolor{gpe}{HTML}{9B4B3B}
\definecolor{loc}{HTML}{3E705C }
\definecolor{per}{RGB}{0,0,0}
\definecolor{org}{RGB}{0,0,0}

\defcitealias{wang-etal-2021-nested}{W21}
\defcitealias{yang-tu-2022-bottom}{Y22}
\defcitealias{shibuya-hovy-2020-nested}{S20}
\defcitealias{corro-2023-dynamic}{C23}
\defcitealias{fu2021nested}{F21}
\defcitealias{lou-etal-2022-nested}{L22}
\defcitealias{shen-etal-2021-locate}{S21}

\title{Nested Named Entity Recognition as Single-Pass Sequence Labeling}

\author{
  \begin{tabular}{cccc}
    \textbf{Alberto Muñoz‐Ortiz,}\textsuperscript{1} &
    \textbf{David Vilares,}\textsuperscript{1} &
    \textbf{Caio Corro,}\textsuperscript{2} &
    \textbf{Carlos Gómez‐Rodríguez}\textsuperscript{1} \\
  \end{tabular} \\[1ex]
  \textsuperscript{1} Universidade da Coruña, CITIC, Spain \\[0.3ex]
  \textsuperscript{2} INSA Rennes, IRISA, Inria, CNRS, Université de Rennes, France \\[0.3ex]
  \texttt{\{alberto.munoz.ortiz,\,david.vilares,\,carlos.gomez\}@udc.es}\\[0.3ex]
  \texttt{caio.corro@insa‐rennes.fr}
}

\newcommand{\wleftarrow}{\rotatebox[origin=c]{-45}{$\leftarrow$}}
\newcommand{\Fleftarrow}{\rotatebox[origin=c]{-45}{$\Leftarrow$}}
\newcommand{\wrightarrow}{\rotatebox[origin=c]{45}{$\rightarrow$}}
\newcommand{\Frightarrow}{\rotatebox[origin=c]{45}{$\Rightarrow$}}

\begin{document}
\maketitle
\begin{abstract}
We cast nested named entity recognition (NNER) as a sequence labeling task by leveraging prior work that linearizes constituency structures, effectively reducing the complexity of this structured prediction problem to straightforward token classification. By combining these constituency linearizations with pretrained encoders, our method captures nested entities while performing exactly $n$ tagging actions. Our approach achieves competitive performance compared to less efficient systems, and it can be trained using any off-the-shelf sequence labeling library.

\end{abstract}

\section{Introduction}
Named Entity Recognition (NER) involves identifying token spans that refer to specific named entities. Traditional approaches use the BIO scheme \cite{ramshaw-marcus-1995-text}, which assigns labels to define the beginning (B), inside (I), and outside (O) tokens of each entity. BIO NER---and other BIO tasks---can be naturally addressed using sequence labeling approaches \cite{ratnaparkhi-1996-maximum,lafferty2001crf,lample-etal-2016-neural,yang-zhang-2018-ncrf}. However, nested NER (NNER) challenges these approaches, as it involves entities embedded within other entities as long as their spans do not overlap. For example, in Figure~\ref{fig:nner}, the person mention ``\textit{president of the USA}'' contains the geopolitical named entity ``\textit{the USA}''. This nested structure breaks the BIO tagging assumption that each token belongs to only one entity. As a result, structured prediction is a better fit for NNER, as it can model interdependent spans and capture hierarchical relationships between nested entities.

\begin{figure}[ht]
    \centering
    \begin{adjustbox}{trim={0.25cm 0cm 0cm 0cm}, clip}
    \begin{tikzpicture}[every node/.style={anchor=base}]

        \begin{scope}[
            level distance=7mm, 
            sibling distance=10mm, 
            yshift=52mm, xshift=-12.5mm,
            edge from parent/.style={draw, decorate, decoration={mirror,amplitude=2pt}}
        ]
        \scriptsize
        \node (tree) [xshift=-10mm] {\texttt{S}}
            child {
                node [font=\ttfamily,text=per] {\texttt{PER}}
                child { node {Lincoln} }
            }
            child { node {was} }
            child {
                node [font=\ttfamily,text=per] {\texttt{PER}}
                child { node {president} }
                child { node {of} }
                child {
                    node [font=\ttfamily,text=org] {\texttt{GPE}}
                    child { node {the} } 
                    child { node {USA} }
                }
            };
        \end{scope}
        \begin{scope}[
            yshift=52mm, xshift=15mm,
            every node/.style={anchor=base},
            level distance=6mm,
            edge from parent/.style={draw, decorate, decoration={mirror, amplitude=2pt}},
            level 1/.style={sibling distance=10mm},
            level 2/.style={sibling distance=6mm},
            level 3/.style={sibling distance=8mm},
            level 4/.style={sibling distance=6mm}
        ] 
        \scriptsize
        \node {\texttt{S}}
        child { 
            node [font=\ttfamily] {\texttt{PER}}
            child { node {Lincoln} }
        }
        child { 
            node [font=\ttfamily] {\texttt{S'}}
            child { node {was} }
            child { 
                node [font=\ttfamily] {\texttt{PER}}
                child { node {president} }
                child { 
                    node [font=\ttfamily] {\texttt{PER'}}
                    child { node {of} }
                    child { 
                        node [font=\ttfamily] {\texttt{GPE}}
                        child { node {the} }
                        child { node {USA} }
                    }
                }
            }
        };
        \end{scope}

        \matrix (m) [
            matrix of nodes,
            row sep=0.5mm,
            column sep=-0.22cm,
            row 1/.style={nodes={font=\small}},
            row 2/.style={nodes={font=\tiny,text depth=1pt}},
            row 3/.style={nodes={font=\tiny,text depth=1pt}},
            row 4/.style={nodes={font=\tiny,text depth=1pt}},
            row 5/.style={nodes={font=\tiny,text depth=1pt}}
        ]
        {
             & Lincoln & was & president & of & the & USA \\[60mm]
        \texttt{abs}
             & (\textbf{\texttt{1}},~\texttt{S},~\texttt{PER})
             & (\textbf{\texttt{1}},~\texttt{S},~–)
             & (\textbf{\texttt{2}},~\texttt{PER},~–)
             & (\textbf{\texttt{2}},~\texttt{PER},~–)
             & (\textbf{\texttt{3}},~\texttt{GPE},~–)
             & (\textbf{\texttt{1}},~\texttt{S},~–) \\
        \texttt{rel}
             & (\textbf{\texttt{1}},~\texttt{S},~\texttt{PER})
             & (\textbf{\texttt{0}},~\texttt{S},~–)
             & (\textbf{\texttt{1}},~\texttt{PER},~–)
             & (\textbf{\texttt{0}},~\texttt{PER},~–)
             & (\textbf{\texttt{1}},~\texttt{GPE},~–)
             & (-\textbf{\texttt{2}},~\texttt{S},~–) \\
        \texttt{dyn}
             & (\textbf{\texttt{1}},~\texttt{S},~\texttt{PER})
             & (\textbf{\texttt{0}},~\texttt{S},~–)
             & (\textbf{\texttt{1}},~\texttt{PER},~–)
             & (\textbf{\texttt{0}},~\texttt{PER},~–)
             & (\textbf{\texttt{1}},~\texttt{GPE},~–)
             & (\textbf{\texttt{1}},~\texttt{S},~–) \\
        \texttt{4tg}
             & (\texttt{\wrightarrow\Frightarrow},~\texttt{S},~\texttt{PER})
             & (\texttt{\wrightarrow\Fleftarrow},~\texttt{S'},~–)
             & (\texttt{\wrightarrow\Fleftarrow},~\texttt{PER},~–)
             & (\texttt{\wrightarrow\Fleftarrow},~\texttt{PER'},~–)
             & (\texttt{\wrightarrow\Fleftarrow},~\texttt{GPE},~–)
             & (\texttt{\wleftarrow},\texttt{GPE},–) \\
        };

        \draw[decorate,decoration={brace,mirror,amplitude=4pt},thick,draw=per]
            ([yshift=-5pt]m-1-2.base west) -- ([yshift=-5pt]m-1-2.base east)
            node[midway,below=4pt,text=per]{\texttt{PER}};
        \draw[decorate,decoration={brace,mirror,amplitude=4pt},thick,draw=per]
            ([yshift=-25pt]m-1-4.base west) -- ([yshift=-25pt]m-1-7.base east)
            node[midway,below=4pt,text=per]{\texttt{PER}};
        \draw[decorate,decoration={brace,mirror,amplitude=4pt},thick,draw=org]
            ([yshift=-5pt]m-1-6.base west) -- ([yshift=-5pt]m-1-7.base east)
            node[midway,below=4pt,text=org]{\texttt{GPE}};
            
        \node[font=\small] (a_label) at ([yshift=89mm]m.south) {(a) Span annotations};
        \node[font=\small] (b_label) at ([yshift=60mm]m.south) {(b) Constituent tree representation and binarized tree};
        \node[font=\small](c_label) at ([yshift=24mm]m.south) {(c) Linearizations};
    \end{tikzpicture}
    \end{adjustbox}
    \caption{
        Illustration of a nested NER annotated sentence as (a) spans, (b) a constituent tree and its corresponding binarized tree, and (c) four ways to linearize that tree. We compare three depth-based---absolute, relative, dynamic---and one transition-based encoding---tetra-tagging (see §\ref{ssec:linearizations}). Binarized non-terminals in the tetra-tagging encoding are marked with '.
    }
    \label{fig:nner}
\end{figure}

Prior work has tackled NNER as a hierarchical or layered sequence labeling task, applying token-level tagging multiple times, either detecting outer entities first and then inner ones \cite{shibuya-hovy-2020-nested}, or identifying inner entities first and then expanding outward \cite{wang-etal-2021-nested}. Other paradigms classify each possible text span as either an entity or not \cite{yu-etal-2020-named,sohrab-miwa-2018-deep,yuan-etal-2022-fusing,corro-2023-dynamic}, use sequence-to-sequence techniques \cite{yan-etal-2021-unified-generative,tan2021sequence}, or exploit hypergraph representations \cite{lu-roth-2015-joint,katiyar-cardie-2018-nested,yan2023nested}.

Alternatively, some methods transform NNER into constituency parsing by mapping samples to constituent trees and relying on parsing algorithms \cite{finkel-manning-2009-nested,wang-etal-2018-neural-transition,fu2021nested,lou-etal-2022-nested}. More recently, \citet{yang-tu-2022-bottom} applied a pointer-network-based bottom-up constituency parser to NNER, offering linear-time decoding. In this context, constituency tree linearization---first introduced by \citet{gomez-rodriguez-vilares-2018-constituent} and later explored in related work \cite{kitaev-klein-2020-tetra,amini-cotterell-2022-parsing}---reframes the structured prediction task as a sequence labeling problem. This approach achieves competitive performance while improving computational efficiency, as it outputs the tree in exactly $n$ tagging actions. This is in contrast with previous sequence-to-sequence models for NNER where the number of tagging actions depends on the number of predicted mentions \cite{miculicich-henderson-2020-partially}.

Building on this line of work, we present a simple yet effective NNER approach that applies constituency tree linearizations within a standard sequence labeling framework.\footnote{Code is available at \url{https://github.com/amunozo/nner_as_sl}.} Our method is single-pass, parser-free, and integrates seamlessly with pretrained encoders. It captures nested structures efficiently and achieves competitive benchmark results, all while requiring minimal implementation effort. Overall, the data preparation remains lightweight and fully automated, and it is no more complex than in traditional nested NER settings, thus preserving the simplicity of standard NER pipelines while adding structural expressiveness.

\section{NNER Through Constituent Parsing as Sequence Labeling}

We represent NNER structures using constituent trees, adopting a sequence labeling strategy 
for syntactic parsing. We hypothesize that this formulation aligns well with NNER due to two key properties: (1) the trees are relatively shallow, involving fewer output levels than in full parsing,\footnote{In full constituent parsing, large gaps between nonterminals have been shown to challenge sequence tagging approaches \cite{vilares-etal-2019-better}.} 
and (2) the encodings provide full coverage of well-formed trees, regardless of nesting depth. 
Furthermore, unlike TreeCRF models \cite{fu2021nested}, span-based \cite{lou-etal-2022-nested} or transition-based parsers \cite{gao2023ntam}, our method avoids complex decoding, enabling linear-time inference\footnote{Although the transformer architecture has quadratic complexity, we follow common practice and refer to the rest of the system’s complexity \cite{shibuya-hovy-2020-nested,corro-2023-dynamic,lou-etal-2022-nested}.}---$n$ tagging actions---with standard architectures.

{
\renewcommand{\arraystretch}{0.8}
\begin{table}[t]
    \centering
    \small
    \begin{tabular}{llccrr}
        \toprule
        \textbf{Dataset} & \textbf{Enc.} 
          & \shortstack{\textbf{\# Labels}\\$(n,c,u)$} 
          & \shortstack{\textbf{Missing}\\$(n,c,u)$} 
          & \textbf{\% dev} & \textbf{\% test} \\
        \midrule
        \multirow{4}{*}{\textbf{ACE2004}} 
          & \texttt{abs} & (6,11,8)   & (0,1,0)  & 100.00 &  99.96 \\
          & \texttt{dyn} & (10,11,8)  & (0,1,0)  & 100.00 &  99.96 \\
          & \texttt{rel} & (10,11,8)  & (0,1,0)  & 100.00 &  99.96 \\
          & \texttt{4tg} & (8,22,8)   & (1,6,1)  & 100.00 &  99.96 \\
        \midrule
        \multirow{4}{*}{\textbf{ACE2005}} 
          & \texttt{abs} & (7,29,9)   & (0,3,1) & 99.88 & 99.98 \\
          & \texttt{dyn} & (9,29,9)   & (0,3,1) & 99.88 & 99.98 \\
          & \texttt{rel} & (10,29,9)  & (0,3,1) & 99.88 & 99.98 \\
          & \texttt{4tg} & (8,58,9)   & (1,6,1) & 99.88 & 99.98 \\
        \midrule
        \multirow{4}{*}{\textbf{GENIA}}   
          & \texttt{abs} & (4,12,13)  & (0,0,0)   & 100.00 & 100.00 \\
          & \texttt{dyn} & (8,12,13)  & (0,0,0)   & 100.00 & 100.00 \\
          & \texttt{rel} & (8,12,13)  & (0,0,0)   & 100.00 & 100.00 \\
          & \texttt{4tg} & (5,22,13)  & (0,1,0) & >99.99 & >99.99 \\
        \midrule
        \multirow{4}{*}{\textbf{NNE}}     
          & \texttt{abs} & (7,264,401)  & (0,14,37) & 99.93 & 99.95 \\
          & \texttt{dyn} & (11,264,401) & (0,14,37) & 99.93 & 99.95 \\
          & \texttt{rel} & (10,264,401) & (0,14,37) & 99.93 & 99.95 \\
          & \texttt{4tg} & (13,399,401) & (0,18,37) & 99.93 & 99.95\\
        \bottomrule
    \end{tabular}
    \caption{Coverage statistics across datasets and encoding strategies. \# Labels indicates the number of unique labels in the training set. Missing shows the number of labels in dev/test not seen during training. \% dev/test indicates the percentage of dev/test labels covered by the training vocabulary.}
    \label{tab:label_vocabulary}
\end{table}
}

\subsection{Linearizations}\label{ssec:linearizations}
We use three depth-based (absolute, relative, dynamic) and one transition-based (tetra-tagging) encoding strategies. These are illustrated in Figure~\ref{fig:nner} using a sentence with nested entities.

Let $\mathbf{w} = [w_1, w_2, \dots, w_{|\mathbf{w}|}]$ be an input word sequence, where $w_i$ represents the word at position $i$. We define a label $l_i = (n_i, c_i, u_i)$ for each word $w_i$,
where:
\begin{enumerate*}[label=(\arabic*)]
\item $n_i$ is an integer indicating the number of common ancestors between words $w_i$ and $w_{i+1}$,
\item $c_i$ is the non-terminal symbol at their lowest common ancestor, and
\item $u_i$ denotes the unary branch for $w_i$, if one exists.
\end{enumerate*}

In the \textbf{absolute encoding (\texttt{abs})}~\citep{gomez-rodriguez-vilares-2018-constituent}, $n^{\text{abs}}_i$ represents the number of common ancestors between $w_i$ and $w_{i+1}$. While \texttt{abs} often suffers from a large and sparse label space in full constituent parsing due to the depth and complexity of syntactic trees, we expect it to remain compact and easier to learn in NNER, where structures are shallower.


The \textbf{relative encoding (\texttt{rel})} was also introduced in~\citep{gomez-rodriguez-vilares-2018-constituent}.
In order to reduce the label set, $n^{\text{rel}}_i$ represents the difference in the number of common ancestors, i.e., for $i > 1,$ $n^{\text{rel}}_i = n^{\text{abs}}_i - n^{\text{abs}}_{i-1}$ (and $n^{\text{rel}}_1 =n^{\text{abs}}_1$).

Combining these two strategies, the \textbf{dynamic encoding (\texttt{dyn})}~\citep{vilares-etal-2019-better} selects, at each position $i$, the most suitable encoding—absolute or relative—for representing the relation between tokens $w_i$ and $w_{i+1}$. It defaults to the relative encoding and switches to absolute when two conditions are met: (1) the relative value $n^{\text{rel}}_i \leq -2$, indicating a sharp drop in tree depth with empty non-terminal levels that must be filled in subsequent steps; and (2) the corresponding absolute value $n^{\text{abs}}_i \leq 3$, ensuring the constituent node is close to the topmost tree level and the absolute label remains compact, avoiding sparsity and improving learnability. The dynamic encoding uses the absolute function only after abrupt structural changes. This affects few labels—0.75\% in ACE2004, 0.73\% in ACE2005, 0.15\% in GENIA, and 1.16\% in NNE—but many sentences: 14.64\%, 12.45\%, 3.86\%, and 20.20\%, respectively.

In addition to depth-based encodings, we also consider the \textbf{tetra-tagging (\texttt{4tg})} linearization introduced by \citet{kitaev-klein-2020-tetra}. Tetra-tagging is a left-corner transition-based parsing algorithm for binary trees, which can be adapted to sequence labeling. It assigns two tags per word: one for the word itself and one for the adjacent fencepost (i.e., the boundary between consecutive words). Each tag encodes whether the word (\wleftarrow,\wrightarrow) or the lowest common ancestor spanning the fencepost (\Fleftarrow$,$\Frightarrow) is a left or right child. To fit our sequence labeling framework, we pair each word and fencepost tag into a single label. We reuse the notation $l_i$, but in this case $n_i$ is refactored to store the two tags associated to $w_i$. Nonterminal symbols and unary chains are added aside the tags as in depth-based encodings.

\paragraph{Label Vocabulary} As entity spans are encoded with a finite set of labels observed during training, the model may face unseen labels at test time. Still, as shown in Table \ref{tab:label_vocabulary}, the proposed encodings cover almost all entities in the development and test sets.

\section{Experiments}
We present our experimental setup and results.

\paragraph{Setup} We use the MaChAmp framework \cite{van-der-goot-etal-2021-massive} to train a multitask transformer-based model.
Specifically, we use \texttt{roberta-large} \cite{liu2019roberta} and \texttt{bert-large-uncased} \cite{devlin-etal-2019-bert} as shared encoders for the ACE and NNE datasets, and \texttt{biobert-large-cased-v1.1} \cite{lee2020biobert} for GENIA. We include RoBERTa as a high-performing option, and BERT for comparison, since it is the most commonly used model in related work. We add three task-specific linear classification heads to predict each atomic component of the label $(n_i, c_i, u_i)$. We adopt the default hyperparameters provided in the MaChAmp repository, as our primary goal is to show that these models perform well in off-the-shelf sequence labeling setups with simple plug-and-play scripts. In particular, for encoding and decoding, we use the scripts provided by the CoDeLin library~\citep{RocVilGomXoveTIC2022}. This library provides homogeneous post-processing to produce trees from ill-formed label sequences; more details about the post-processing are in Appendix~\ref{app:post-processing}.

\begin{table}[t]
    \centering
    \small
    \begin{tabular}{lcrrrr}
        \toprule
        \textbf{Model} & \textbf{Comp.} & \textbf{ACE2004} & \textbf{ACE2005} & \textbf{GENIA} & \textbf{NNE} \\
        \midrule
        \multicolumn{6}{l}{\textbf{Other approaches} ($O(n)$)} \\
        \midrule
        \citetalias{wang-etal-2021-nested} & $O(n)$ & $86.1_{\phantom{\pm0.3}}$ & $84.7_{\phantom{\pm0.3}}$ & $\mathbf{78.7_{\phantom{\pm0.3}}}$ & –$_{\phantom{\pm0.3}}$ \\
        \citetalias{yang-tu-2022-bottom} & $O(n)$ & $86.9_{\phantom{\pm0.3}}$ & $85.5_{\phantom{\pm0.3}}$ & $78.2_{\phantom{\pm0.3}}$ & –$_{\phantom{\pm0.3}}$ \\
        \midrule
        \multicolumn{6}{l}{\textbf{Other approaches} (higher complexity)} \\
        \midrule
        \citetalias{shibuya-hovy-2020-nested} & $O(n^2)$ & $85.8_{\phantom{\pm0.3}}$ & $84.3_{\phantom{\pm0.3}}$ & $77.4_{\phantom{\pm0.3}}$ & –$_{\phantom{\pm0.3}}$ \\
        \citetalias{corro-2023-dynamic} & $O(n^2)$ & $86.2_{\phantom{\pm0.3}}$ & $84.8_{\phantom{\pm0.3}}$ & $78.3_{\phantom{\pm0.3}}$ & –$_{\phantom{\pm0.3}}$ \\
        \citetalias{shen-etal-2021-locate} & $O(n^2)$ & $87.4_{\phantom{\pm0.3}}$ & $86.7_{\phantom{\pm0.3}}$  & $80.5_{\phantom{\pm0.3}}$ &  –$_{\phantom{\pm0.3}}$ \\
        \citetalias{fu2021nested} & $O(n^3)$ & $86.6_{\pm0.3}$ & $85.4_{\pm0.1}$ & $78.2_{\pm0.1}$ & –$_{\phantom{\pm0.3}}$ \\
        \citetalias{lou-etal-2022-nested} & $O(n^4)$ & $87.9_{\phantom{\pm0.3}}$ & $86.9_{\phantom{\pm0.3}}$ & $78.4_{\phantom{\pm0.3}}$ & $94.6_{\phantom{\pm0.3}}$ \\

        \midrule
        \multicolumn{6}{l}{\textbf{This work} ($O(n)$)} \\
        \midrule 
        \multicolumn{6}{l}{RoBERTa (ACE and NNE) and BioBERT} \\
        \midrule
        \texttt{abs} & $O(n)$ & $86.8_{\pm0.3}$ & $85.0_{\pm0.3}$ & $76.6_{\pm0.3}$ & $94.1_{\pm0.3}$ \\
        \texttt{rel} & $O(n)$ & $86.9_{\pm0.3}$ & $85.5_{\pm0.3}$ & $75.1_{\pm0.7}$ & $\mathbf{94.3_{\pm0.1}}$ \\
        \texttt{dyn} & $O(n)$ & $\mathbf{87.7_{\pm0.2}}$ & $\mathbf{86.0_{\pm0.3}}$ & $75.3_{\pm0.5}$ & $\mathbf{94.3_{\pm0.2}}$ \\
        \texttt{4tg} & $O(n)$ & $86.1_{\pm0.5}$ & $84.7_{\pm0.5}$ & $74.6_{\pm0.8}$ & $94.1_{\pm0.1}$ \\
        \midrule
        \multicolumn{6}{l}{BERT (for comparison)} \\
        \midrule
        \texttt{abs} & $O(n)$ & $86.1_{\pm0.3}$ & $83.8_{\pm0.4}$ & –$_{\phantom{\pm0.3}}$ & $94.0_{\pm0.1}$ \\
        \texttt{rel} & $O(n)$ & $87.0_{\pm0.4}$ & $85.2_{\pm0.3}$ & –$_{\phantom{\pm0.3}}$ & $94.1_{\pm0.1}$ \\
        \texttt{dyn} & $O(n)$ & $87.0_{\pm0.1}$ & $85.2_{\pm0.5}$ & –$_{\phantom{\pm0.3}}$ & $94.2_{\pm0.0}$ \\
        \texttt{4tg} & $O(n)$ & $86.0_{\pm0.5}$ & $84.7_{\pm0.2}$ & –$_{\phantom{\pm0.3}}$ & $94.1_{\pm0.1}$ \\
        \bottomrule
    \end{tabular}
    \caption{
    F1-scores for different datasets and encodings, a comparison to prior work, and computational complexities. Highest F1 among $O(n)$ models is in bold.}
    \label{tab:results}
\end{table}

\begin{table}[hbpt]
    \centering
    \small
    \begin{tabular}{lc cc cc cc cc}
        \toprule
         &  & \multicolumn{2}{c}{\textbf{ACE2004}} & \multicolumn{2}{c}{\textbf{ACE2005}} & \multicolumn{2}{c}{\textbf{GENIA}} & \multicolumn{2}{c}{\textbf{NNE}} \\
        \cmidrule(lr){3-4} \cmidrule(lr){5-6} \cmidrule(lr){7-8} \cmidrule(lr){9-10}
         & & \textbf{Prec.} & \textbf{Rec.} & \textbf{Prec.} & \textbf{Rec.} & \textbf{Prec.} & \textbf{Rec.} & \textbf{Prec.} & \textbf{Rec.} \\
        \midrule
        \multirow{5}{*}{\texttt{abs}}
              & 0    & \textbf{84.4} & \textbf{87.2} & \textbf{83.1} & \textbf{86.7} & 78.9 & \textbf{79.8} & 88.8         & \textbf{92.3} \\
              & 1    & 78.9          & 85.9          & 71.0          & 84.3          & 44.4          & 43.5          & 90.6         & 95.4          \\
              & $\ge$1 & \textbf{77.7}          & \textbf{74.6}          & 69.8          & 67.1          & 44.5          & 37.5          & 88.8         & 88.4          \\
              & $\ge$2 & \textbf{71.6} & 63.6          & 61.3 & 37.4          & 50.0 & \textbf{25.0} & 85.2         & 83.2          \\
        \midrule
        \multirow{5}{*}{\texttt{rel}}
              & 0    & 83.5          & 85.8          & 81.6          & 85.9          & 77.7          & 76.1          & 87.8         & 91.7          \\
              & 1    & 78.7          & 87.6          & 72.7          & 85.8          & \textbf{58.1} & 41.7          & 90.1         & 95.4          \\
              & $\ge$1 & 75.9          & 72.6          & 71.1          & 68.8          & \textbf{57.4}          & 23.5          & \textbf{88.9}         & 87.3          \\
              & $\ge$2 & 62.7          & 62.3          & 59.6          & 40.5          & 0.0           & 0.0           & \textbf{86.4} & 82.4          \\
        \midrule
        \multirow{5}{*}{\texttt{dyn}}
              & 0    & 84.0          & 86.8          & 82.6          & 86.5          & 78.1          & 76.3          & 88.1         & 91.9          \\
              & 1    & \textbf{79.4} & \textbf{88.1} & \textbf{75.3} & \textbf{86.9} & 51.7          & 42.9          & 90.4         & \textbf{95.6} \\
              & $\ge$1 & 77.4          & 72.1          & 72.7          & 70.5          & 51.6          & 28.3          & 88.7         & 87.7          \\
              & $\ge$2 & 68.0          & \textbf{66.2} & 58.6          & 49.6          & 0.0           & 0.0           & 85.3         & 82.7          \\
        \midrule
        \multirow{5}{*}{\texttt{4tg}}
              & 0    & 83.1          & 86.0          & 81.2          & 85.7          & 78.2          & 77.8          & \textbf{89.3} & 92.2          \\
              & 1    & 73.6          & 86.8          & 73.4          & 85.2          & 32.5          & \textbf{45.6} & \textbf{90.8} & 95.5          \\
              & $\ge$1 & 70.0          & 72.9          & 70.8          & \textbf{71.0}          & 30.5          & 36.2          & 88.2         & \textbf{88.7}          \\
              & $\ge$2 & 53.9          & 59.6          & 59.2          & \textbf{58.8} & 3.8           & 10.0          & 83.5         & \textbf{84.9} \\
        \midrule
        \midrule
        \multirow{5}{*}{\citetalias{shibuya-hovy-2020-nested}}
              & 0    & --   & --   & 80.4 & 76.1 & \textbf{80.3} & 77.9    & --   & --   \\
              & 1    & --   & --   & 72.4 & 71.7 & 57.1 & 40.6 & --   & --   \\
              & $\ge$1 & -- & -- & \textbf{73.2} & 70.4 & 57.1 & \textbf{40.6} & -- & --     \\
              & $\ge$2 & --   & --   & \textbf{77.6} & 57.5 & \textbf{66.7} & 0.0    & --   & --   \\
        \bottomrule
    \end{tabular}
    \caption{Precision and recall by encoding and depth per dataset for entities at different depths. Best values for each dataset and depth shown in bold. Results for \citet{shibuya-hovy-2020-nested} are added from their published scores.}
    \label{tab:results_depth}
\end{table}

\paragraph{Datasets} We train and evaluate our models on four popular English NNER benchmarks: GENIA \cite{kim2003genia}, ACE 2004 \cite{doddington-etal-2004-automatic}, ACE 2005 \cite{walker2006ace}, and NNE \cite{ringland2019nne}. For ACE and GENIA, we preprocess the datasets following \citet{shibuya-hovy-2020-nested}. Information about the datasets is shown in Appendix Table \ref{tab:datasets}. Notably, most entities are flat and short and highly nested spans are rare.

\paragraph{Metrics} We evaluate our models in terms of precision, recall, and F1-scores based on strict entity boundaries, requiring both correct spans and types. We also provide finer-grained evaluations (e.g., by entity length) and external comparisons.
\subsection{Results}
Table~\ref{tab:results} reports precision, recall, and F1-scores for our sequence labeling models compared to previous work on NNER. All results are averaged over five runs to provide robust performance estimates (standard deviations are also reported).

Our approach outperforms both comparable and more complex methods on the ACE and NNE datasets. However, this is not the case for GENIA, where the results---particularly recall---are lower. Although the \texttt{dyn} encoding predominantly uses the relative function—as most entities are flat or shallow---it outperforms the other encodings on every dataset except GENIA, effectively leveraging the best of \texttt{abs} and \texttt{rel} linearizations. Even though the absolute function is rarely used in the dynamic encoding, it yields small but consistent F1 improvements of 0.74\% for ACE2004, 0.39\% for ACE2005, 0.22\% for GENIA, and 0.02\% for NNE. As more than 99\% of entities in GENIA are flat or singly nested, the \texttt{dyn} and \texttt{rel} encodings represent the entities almost identically. Tetra-tagging is the worst-performing encoding in all but one setup.

To better understand how linearizations handle nested entities, we evaluate each encoding by analyzing model performance across different entity depth and length, using RoBERTa and BioBERT.

\begin{table}[t!]
    \centering
    \small
    \begin{tabular}{lc cc cc cc cc}
        \toprule
         &  & \multicolumn{2}{c}{\textbf{ACE2004}} & \multicolumn{2}{c}{\textbf{ACE2005}} & \multicolumn{2}{c}{\textbf{GENIA}} & \multicolumn{2}{c}{\textbf{NNE}} \\
        \cmidrule(lr){3-4} \cmidrule(lr){5-6} \cmidrule(lr){7-8} \cmidrule(lr){9-10}
         & & \textbf{Prec.} & \textbf{Rec.} & \textbf{Prec.} & \textbf{Rec.} & \textbf{Prec.} & \textbf{Rec.} & \textbf{Prec.} & \textbf{Rec.} \\
        \midrule
        \multirow{6}{*}{\texttt{abs}}
            & 1       & 84.8  & 86.4  & 83.6  & \textbf{86.4} & 78.4  & \textbf{77.9}  & 93.2  & 93.2  \\
            & 2--4    & 88.1  & \textbf{87.1}  & 79.2  & 82.2         & \textbf{74.8}  & \textbf{73.9}  & 89.8  & 90.3  \\
            & $\geq$2      & 83.8 & 83.6 & 77.2 & 79.1 & \textbf{74.3} & \textbf{74.2} & 89.5 & 88.9 \\
            & 5--9    & 76.8  & 78.2  & 74.1  & \textbf{76.8}  & \textbf{69.3}  & \textbf{76.7}  & 85.0  & \textbf{73.6}  \\
            & $\ge$10 & \textbf{70.5}  & \textbf{71.0}  & 67.9  & 61.5         & \textbf{81.8}  & \textbf{75.0}  & \textbf{80.0}  & 20.0  \\
        \midrule
        \multirow{6}{*}{\texttt{rel}}
            & 1       & \textbf{89.0}  & 88.0         & \textbf{86.9}  & 85.1         & 79.5          & 76.6          & \textbf{94.2}  & 93.4  \\
            & $\geq$2      & 83.7 & 80.3 & 79.4 & 78.0 & 73.4 & 68.1 & \textbf{90.6} & 87.8 \\
            & 2--4    & 89.5          & 86.3         & \textbf{82.5}  & 82.4         & 74.1          & 68.3          & \textbf{91.0}  & 89.6  \\
            & 5--9    & 77.8  & 73.2         & 75.8          & 70.8         & 68.2          & 67.7          & \textbf{85.5}  & 68.3  \\
            & $\ge$10 & 55.3          & 53.4         & 63.7          & 60.9         & 65.3          & 53.3          & 44.4          & 20.0  \\
        \midrule
        \multirow{6}{*}{\texttt{dyn}}
            & 1       & 88.9          & 88.2 & 86.4          & 86.2         & \textbf{79.8}  & 77.2          & 94.0          & 93.4  \\
            & $\geq$2      & 84.6 & 82.2 & \textbf{80.0} & \textbf{79.3} & 73.4 & 66.8 & 90.3 & 88.1 \\
            & 2--4    & \textbf{89.8}  & 86.7         & 82.4          & \textbf{83.0} & 73.9          & 66.8          & 90.7          & 89.8  \\
            & 5--9    & 77.4          & 75.4         & \textbf{76.7}  & 72.6         & 69.2          & 67.7          & 84.4          & 69.9  \\
            & $\ge$10 & 64.5          & 65.1         & \textbf{69.3}  & \textbf{67.1} & 65.2          & 50.0          & 50.0          & 20.0  \\            
        \midrule
        \multirow{6}{*}{\texttt{4tg}}
            & 1       & 88.3          & 87.6         & 86.1          & 85.8         & 78.7          & 77.8          & 93.9          & \textbf{93.5}  \\
            & $\geq$2      & 79.9 & 80.4 & 75.3 & 77.0 & 69.0 & 70.2 & 88.8 & \textbf{89.3} \\
            & 2--4    & 87.8          & 85.5         & 79.8          & 80.8         & 72.2          & 70.4          & 89.9          & \textbf{91.0}  \\
            & 5--9    & 72.6          & 74.6         & 72.2          & 71.2         & 56.7          & 68.9          & 76.9          & 72.0  \\
            & $\ge$10 & 48.3          & 57.8         & 52.6          & 62.1         & 16.8          & 58.3          & 18.2          & \textbf{30.0}  \\            
        \midrule
        \midrule
        \multirow{6}{*}{\citetalias{shen-etal-2021-locate}}
            & 1       & \textbf{89.6}  & \textbf{91.0}  & --    & --    & --    & --    & --    & --    \\
            & $\geq$2      & \textbf{85.3} & \textbf{83.7} & --    & --    & --    & --    & --    & --  \\
            & 2--4    & 87.0  & 86.0  & --    & --    & --    & --    & --    & --    \\
            & 5--9    & \textbf{84.3}  & \textbf{83.2}  & --    & --    & --    & --    & --    & --    \\
            & $\ge$10 & 69.2  & 63.2  & --    & --    & --    & --    & --    & --    \\
        \bottomrule
    \end{tabular}
    \caption{Precision and recall by encoding and length for entities of different span lengths.
    Best values for each dataset and span length shown in bold. Results from \citet{shen-etal-2021-locate} are added for comparison.}
    \label{tab:results_length}
\end{table}

\begin{table*}[hpbt]
    \centering
    \footnotesize
    \begin{tabular}{l cccc cccc cccc cccc}
        \toprule
        & \multicolumn{4}{c}{\textbf{ACE2004}} & \multicolumn{4}{c}{\textbf{ACE2005}} & \multicolumn{4}{c}{\textbf{GENIA}} & \multicolumn{4}{c}{\textbf{NNE}} \\
        \cmidrule(lr){2-5} \cmidrule(lr){6-9} \cmidrule(lr){10-13} \cmidrule(lr){14-17}
         & \textbf{Best} & \textbf{Worst} & \textbf{Mean} & \textbf{Std} & \textbf{Best} & \textbf{Worst} & \textbf{Mean} & \textbf{Std} & \textbf{Best} & \textbf{Worst} & \textbf{Mean} & \textbf{Std} & \textbf{Best} & \textbf{Worst} & \textbf{Mean} & \textbf{Std} \\
        \midrule
        \texttt{4tg} & \texttt{PER} & \texttt{FAC} & 78.9 & 10.6 & \texttt{PER} & \texttt{LOC} & 77.8 & 7.4 & \texttt{protein} & \texttt{DNA} & 73.1 & 3.0 & \texttt{ENERGY} & \texttt{BAND} & 73.7 & 29.8 \\
        
        \texttt{abs} & \texttt{PER} & \texttt{WEA} & 78.5 & 10.8 & \texttt{PER} & \texttt{LOC} & 78.6 & 7.2 & \texttt{protein} & \texttt{cell\_type} & 74.7 & 3.4 & \texttt{DISEASE} & \texttt{BAND} & 74.4 & 29.5 \\
        
        \texttt{dyn} & \texttt{PER} & \texttt{FAC} & 81.3 & 8.3 & \texttt{PER} & \texttt{LOC} & 79.6 & 7.6 & \texttt{protein} & \texttt{DNA} & 73.1 & 3.2 & \texttt{ENERGY} & \texttt{BAND} & 74.0 & 29.4 \\
        
        \texttt{rel} & \texttt{PER} & \texttt{FAC} & 80.6 & 8.8 & \texttt{PER} & \texttt{LOC} & 79.3 & 7.7 & \texttt{protein} & \texttt{DNA} & 73.0 & 3.2 & \texttt{HURRICANE} & \texttt{BAND} & 73.4 & 30.4 \\
        \bottomrule
    \end{tabular}
    \caption{F1-score statistics across datasets and encodings. The table identifies the best and worst performing entity labels and lists the mean and standard deviation of F1-scores across all labels.}
    \label{tab:results_type}
\end{table*}

\paragraph{Results by depth}
Table \ref{tab:results_depth} presents precision and recall by depth for each encoding and dataset. Since the number of predicted entities may vary across runs, we follow \citet{shibuya-hovy-2020-nested} and report the precision results from the run whose F1 is closest to the average, and the average recall over five runs. \texttt{abs} performs best for flat entities on all datasets, \texttt{dyn} leads for singly nested entities in ACE2004 and ACE2005, \texttt{rel} in GENIA, and \texttt{4tg} and \texttt{dyn} tie in NNE. For entities nested two or more times, \texttt{abs} performs best in GENIA and achieves the highest precision on the ACE datasets, but is outperformed in recall except in GENIA. In NNE, \texttt{4tg} achieves the highest recall.
We also compare against external work that has reported these results, particularly \citet{shibuya-hovy-2020-nested}, where our method shows superior performance on flat entities but underperforms on deep nested structures.

\paragraph{Results by entity length}
Table \ref{tab:results_length} presents precision and recall by span length. Among the encodings, \texttt{rel} and \texttt{dyn} tend to perform best on single-token spans and those of 2–4 tokens. Results for 5–9 token spans are mixed, while \texttt{abs} consistently outperforms the others on longer spans ($\geq$10 tokens), except in ACE2005, where it is surpassed by \texttt{dyn}. \texttt{4tg} performs well on short spans—particularly in NNE—but its performance drops sharply as span length increases. These trends suggest that \texttt{dyn} is well-suited for frequent, shorter entities, while \texttt{abs} offers greater robustness on longer ones. 
In addition, we benchmark against external work we are aware of, such as the more complex model from \citet{shen-etal-2021-locate}, where our methods are competitive for entities spanning fewer than four tokens, but the performance gap widens for longer entities where their specialized architecture appears more advantageous.

\paragraph{Results by entity type}

Table \ref{tab:results_type} provides a detailed analysis of F1-scores for each entity type, identifying the best and worst performers and summarizing the overall distribution with mean and standard deviation. \texttt{PER} (Person) consistently achieves the highest scores in both ACE datasets, while \texttt{protein} is the strongest performer in GENIA, regardless of the encoding used.

Focusing on the variance, scores in GENIA and ACE2005 are stable across types, with a standard deviation of around 3 and 7.5 respectively for all encodings. For ACE2004, the \texttt{rel} and \texttt{dyn} encodings show a standard deviation of approximately 8.5, while \texttt{4tg} and \texttt{abs} present slightly higher variation at around 10. As expected, NNE shows a much higher standard deviation ($\sim$29.7), due to its large and sparse label set; however, this pattern is stable across encodings.

\subsection{Discussion}

The results show that with standard transformer-based encoders, NNER can be effectively learned as sequence labeling under various representations. As expected, for both flat and short entities, differences across encodings do not appear to substantially affect performance, since in these cases representing sentences as constituent trees does not lead to significant changes. For deep entities, more elaborate encodings appear to be beneficial compared to the naive absolute encoding. For longer entities, results are mixed. This is likely influenced by the fact that longer entities can be either flat or deep, which may confound conclusions about the impact of different encodings. In particular, depth-based encodings perform better on the ACE and GENIA datasets, while \texttt{4tg} performs comparably to the other encodings on the NNE dataset.
In this context, a factor could be the relative ease of the NNE dataset. Although it includes many nested entities and a large label set, it also provides a substantial larger amount of training data (see Table \ref{tab:datasets}). If the learning task becomes less demanding overall, differences between encodings are less pronounced, and the encoder may play a more prominent role. As shown by our NNE results, the narrow F1 range (94.0–94.6) across encodings and models suggests that, with enough data, different encodings are learned to similar capacity, thereby limiting observable performance differences.

\section{Conclusion}

We cast nested NER for the first time as a single-pass sequence labeling task by first transforming NNER annotations into constituent trees, and then leveraging the linearization options these trees enable. In particular, we explored both depth-based (absolute, relative and dynamic encodings) and transition-based encodings (tetra-tagging) previously proposed for this type of tree structure. Our experiments show that this approach models nested structures effectively, with competitive performance and no need for complex architectures, remaining lightweight and compatible with standard sequence labeling tools.

Among the explored encodings, the dynamic variant---which selects between absolute and relative schemes based on local structural cues---consistently achieves the highest F1 across datasets, except for GENIA. The transition-based tetra-tagging approach is
competitive on flat and short entities but degrades on longer spans, where depth-based strategies prove more effective. 

\section*{Limitations}
Due to the limited availability of freely accessible NNER datasets, we focus on the most widely used English benchmarks, including both public and proprietary ones. These datasets are commonly used in prior work and provide a solid basis for comparison. While our approach is evaluated on these established datasets, it’s important to note that results may vary across benchmarks, as factors such as annotation scheme, entity types, nesting depth, domain, and language can influence performance.

\section*{Acknowledgements}
This work was funded by SCANNER-UDC (PID2020-113230RB-C21) funded by MICIU/AEI/10.13039/501100011033; Xunta de Galicia (ED431C 2024/02); GAP (PID2022-139308OA-I00) funded by MICIU/AEI/10.13039/501100011033/ and by ERDF, EU; Grant PRE2021-097001 funded by MICIU/AEI/10.13039/501100011033 and by ESF+ (predoctoral training grant associated to project PID2020-113230RB-C21); LATCHING (PID2023-147129OB-C21) funded by MICIU/AEI/10.13039/501100011033 and ERDF; TSI-100925-2023-1 funded
by Ministry for Digital Transformation and Civil
Service and “NextGenerationEU” PRTR; 
and Centro de Investigación de Galicia ‘‘CITIC’’, funded by the Xunta de Galicia through the collaboration agreement between the Consellería de Cultura, Educación, Formación Profesional e Universidades and the Galician universities for the reinforcement of the research centres of the Galician University System (CIGUSA). 

This research project was made possible through the access granted by the Galician Supercomputing Center (CESGA) to its supercomputing infrastructure. The supercomputer FinisTerrae III and its permanent data storage system have been funded by the NextGeneration EU 2021 Recovery, Transformation and Resilience Plan, ICT2021-006904, and also from the Pluriregional Operational Programme of Spain 2014-2020 of the European Regional Development Fund (ERDF), ICTS-2019-02-CESGA-3, and from the State Programme for the Promotion of Scientific and Technical Research of Excellence of the State Plan for Scientific and Technical Research and Innovation 2013-2016 State subprogramme for scientific and technical infrastructures and equipment of ERDF, CESG15-DE-3114.

Caio Corro has received funding from the French Agence Nationale pour la Recherche under grant agreement InExtenso ANR-23-IAS1-0004 and SEMIAMOR ANR-23-CE23-0005.

\bibliography{custom, anthology}

\appendix

\section{Precision and Recall Results}

Table~\ref{tab:results-prrec} shows the precision and recall corresponding to the evaluation presented in Table~\ref{tab:results} of the main text.

\begin{table*}[hptb]
    \centering
    \small
    \setlength{\tabcolsep}{1.2pt}
    \begin{tabular}{l c *{4}{cc}}
        \toprule
         & & \multicolumn{2}{c}{\textbf{ACE2004}} & \multicolumn{2}{c}{\textbf{ACE2005}} & \multicolumn{2}{c}{\textbf{GENIA}} & \multicolumn{2}{c}{\textbf{NNE}} \\
         \cmidrule(lr){3-4} \cmidrule(lr){5-6} \cmidrule(lr){7-8} \cmidrule(lr){9-10}
         \textbf{Model} & \textbf{Comp.} & \textbf{Prec.} & \textbf{Rec.} & \textbf{Prec.} & \textbf{Rec.} & \textbf{Prec.} & \textbf{Rec.} & \textbf{Prec.} & \textbf{Rec.} \\
        \midrule
        \multicolumn{10}{l}{\textbf{Other approaches} ($O(n)$)} \\
        \midrule
        \citetalias{wang-etal-2021-nested} & $O(n)$ & $85.4_{\phantom{\pm0.5}}$ & $86.7_{\phantom{\pm0.5}}$ & $84.2_{\phantom{\pm0.5}}$ & $85.3_{\phantom{\pm0.5}}$ & $78.2_{\phantom{\pm0.5}}$ & $\mathbf{79.2_{\phantom{\pm0.5}}}$ & –$_{\phantom{\pm0.5}}$ & –$_{\phantom{\pm0.5}}$ \\
        \citetalias{yang-tu-2022-bottom}  & $O(n)$ & $86.6_{\phantom{\pm0.5}}$ & $87.3_{\phantom{\pm0.5}}$ & $84.6_{\phantom{\pm0.5}}$ & $\mathbf{86.4_{\phantom{\pm0.5}}}$ & $78.1_{\phantom{\pm0.5}}$ & $78.3_{\phantom{\pm0.5}}$ & –$_{\phantom{\pm0.5}}$ & –$_{\phantom{\pm0.5}}$ \\
        \midrule
        \multicolumn{10}{l}{\textbf{Other approaches} (higher complexity)} \\
        \midrule
        \citetalias{shibuya-hovy-2020-nested} & $O(n^2)$ & $85.9_{\phantom{\pm0.5}}$ & $85.7_{\phantom{\pm0.5}}$ & $83.8_{\phantom{\pm0.5}}$ & $84.9_{\phantom{\pm0.5}}$ & $77.8_{\phantom{\pm0.5}}$ & $76.9_{\phantom{\pm0.5}}$ & –$_{\phantom{\pm0.5}}$ & –$_{\phantom{\pm0.5}}$ \\
        \citetalias{corro-2023-dynamic}    & $O(n^2)$ & $87.4_{\phantom{\pm0.5}}$ & $85.0_{\phantom{\pm0.5}}$ & $84.4_{\phantom{\pm0.5}}$ & $85.3_{\phantom{\pm0.5}}$ & $79.3_{\phantom{\pm0.5}}$ & $77.3_{\phantom{\pm0.5}}$ & –$_{\phantom{\pm0.5}}$ & –$_{\phantom{\pm0.5}}$ \\
        \citetalias{shen-etal-2021-locate}  & $O(n^2)$ & $87.4{\phantom{\pm0.5}}$ & $87.4{\phantom{\pm0.5}}$ & $86.1_{\phantom{\pm0.5}}$ & $87.3_{\phantom{\pm0.5}}$ & $80.2_{\phantom{\pm0.5}}$ & $80.9_{\phantom{\pm0.5}}$ & –$_{\phantom{\pm0.5}}$ & –$_{\phantom{\pm0.5}}$ \\
        \citetalias{fu2021nested}          & $O(n^3)$ & $86.7_{\pm0.4}$        & $86.5_{\pm0.4}$        & $84.5_{\pm0.4}$        & $86.4_{\pm0.2}$        & $78.2_{\pm0.7}$        & $78.2_{\pm0.8}$        & –$_{\phantom{\pm0.5}}$ & –$_{\phantom{\pm0.5}}$ \\
        \citetalias{lou-etal-2022-nested}  & $O(n^4)$ & $87.4_{\phantom{\pm0.5}}$ & $88.4_{\phantom{\pm0.5}}$ & $86.0_{\phantom{\pm0.5}}$ & $87.9_{\phantom{\pm0.5}}$ & $78.4_{\phantom{\pm0.5}}$ & $78.5_{\phantom{\pm0.5}}$ & $94.3_{\phantom{\pm0.5}}$ & $95.0_{\phantom{\pm0.5}}$ \\
        \midrule
        \multicolumn{10}{l}{\textbf{This work} ($O(n)$)} \\
        \midrule 
        \multicolumn{10}{l}{RoBERTa (ACE and NNE) and BioBERT} \\
        \midrule
        \texttt{abs} & $O(n)$ & $86.8_{\pm0.5}$        & $86.8_{\pm0.1}$        & $84.1_{\pm0.6}$        & $85.9_{\pm0.4}$        & $77.6_{\pm0.4}$        & $75.6_{\pm0.3}$        & $94.0_{\pm0.3}$        & $94.2_{\pm0.3}$ \\
        \texttt{rel} & $O(n)$ & $87.4_{\pm0.4}$        & $86.4_{\pm0.3}$        & $85.3_{\pm0.6}$        & $85.7_{\pm0.3}$        & $78.4_{\pm0.7}$        & $72.1_{\pm1.8}$        & $\mathbf{94.4_{\pm0.0}}$ & $94.1_{\pm0.1}$ \\
        \texttt{dyn} & $O(n)$ & $\mathbf{88.0_{\pm0.3}}$ & $\mathbf{87.3_{\pm0.2}}$ & $\mathbf{85.6_{\pm0.4}}$ & $\mathbf{86.4_{\pm0.5}}$        & $\mathbf{78.5_{\pm0.5}}$ & $72.5_{\pm1.3}$        & $94.3_{\pm0.2}$        & $94.2_{\pm0.2}$ \\
        \texttt{4tg} & $O(n)$ 
            & $85.8_{\pm1.1}$ & $86.4_{\pm0.1}$ 
            & $84.0_{\pm0.5}$ & $85.5_{\pm0.5}$ 
            & $75.2_{\pm1.1}$ & $74.1_{\pm0.9}$ 
            & $94.0_{\pm0.2}$ & $\mathbf{94.3_{\pm0.1}}$ \\
        \midrule 
        \multicolumn{6}{l}{BERT (for comparison)} \\
        \midrule
        \texttt{abs} & $O(n)$ & $85.9_{\pm0.4}$ & $86.2_{\pm0.5}$ & $82.7_{\pm0.4}$ & $85.0_{\pm0.4}$ & –$_{\phantom{\pm0.5}}$ & –$_{\phantom{\pm0.5}}$ & $94.1_{\pm0.1}$ & $94.0_{\pm0.1}$ \\
        \texttt{rel} & $O(n)$ & $87.9_{\pm0.4}$ & $86.0_{\pm0.4}$ & $84.9_{\pm0.7}$ & $85.5_{\pm0.0}$ & –$_{\phantom{\pm0.5}}$ & –$_{\phantom{\pm0.5}}$ & $\mathbf{94.4_{\pm0.1}}$ &$93.8_{\pm0.1}$ \\
        \texttt{dyn} & $O(n)$ & $87.9_{\pm0.1}$ & $86.2_{\pm0.2}$ & $84.9_{\pm0.5}$ & $85.4_{\pm0.7}$ & –$_{\phantom{\pm0.3}}$ &  –$_{\phantom{\pm0.3}}$ & $\mathbf{94.4_{\pm0.0}}$ & $94.0_{\pm0.1}$ \\
                \texttt{4tg} & $O(n)$ 
            & $86.0_{\pm0.7}$ & $86.0_{\pm0.3}$ 
            & $83.6_{\pm0.2}$ & $85.8_{\pm0.3}$ 
            & –$_{\phantom{\pm0.5}}$ & –$_{\phantom{\pm0.5}}$ 
            & $94.1_{\pm0.1}$ & $94.1_{\pm0.1}$ \\
        \bottomrule
    \end{tabular}
    \caption{Performance metrics for each dataset and encoding. Among $O(n)$ methods, the top values for each dataset are in \textbf{bold}.}
    \label{tab:results-prrec}
\end{table*}

\section{Results from Best Runs}

Table~\ref{tab:best_results} presents the precision, recall, and F1-scores from the best run of each model, selected according to the highest F1-score.

\begin{table*}[t]
    \centering
    \small
    \begin{tabular}{l rrr rrr rrr rrr}
        \toprule
        & \multicolumn{3}{c}{\textbf{ACE2004}} & \multicolumn{3}{c}{\textbf{ACE2005}} & \multicolumn{3}{c}{\textbf{GENIA}} & \multicolumn{3}{c}{\textbf{NNE}} \\
        \cmidrule(lr){2-4} \cmidrule(lr){5-7} \cmidrule(lr){8-10} \cmidrule(lr){11-13}
        \textbf{Model} & \textbf{Prec.} & \textbf{Rec.} & \textbf{F1} & \textbf{Prec.} & \textbf{Rec.} & \textbf{F1} & \textbf{Prec.} & \textbf{Rec.} & \textbf{F1} & \textbf{Prec.} & \textbf{Rec.} & \textbf{F1} \\
        \midrule
        \texttt{abs} & 86.4 & 86.6 & 86.5 & 83.7 & \textbf{85.7} & 84.7 & 77.1 & \textbf{75.5} & \textbf{76.3} & 93.6 & 93.8 & 93.7 \\
        \texttt{rel} & 86.9 & 86.1 & 86.5 & 85.1 & 85.1 & 85.1 & \textbf{79.3} & 69.8 & 74.3 & \textbf{94.3} & 93.9 & \textbf{94.1} \\
        \texttt{dyn} & \textbf{87.7} & \textbf{87.2} & \textbf{87.4} & \textbf{85.2} & \textbf{85.7} & \textbf{85.5} & 78.5 & 71.4 & 74.8 & 93.9 & 93.9 & 93.9 \\
        \texttt{4tg} & 83.8 & 86.6 & 85.1 & 83.1 & 84.7 & 83.9 & 74.6 & 72.6 & 73.5 & 93.7 & \textbf{94.2} & 93.9 \\
        \bottomrule
    \end{tabular}
    \caption{Metrics from each model's best run, selected by F1-score. The highest value for each metric and dataset is highlighted in bold.}
    \label{tab:best_results}
\end{table*}

\section{Dataset information}

Table~\ref{tab:datasets} presents statistics for each dataset, including information on entity depth and length.

\begin{table*}[t]
    \centering
    \small
    \begin{tabular}{lcccc}
        \toprule
        \textbf{Characteristic} & \textbf{ACE2004} & \textbf{ACE2005} & \textbf{GENIA} & \textbf{NNE} \\
        \midrule
        Topic           & News          & News          & Biomedical & News \\
        Sentences       & 7\,762        & 9\,335        & 18\,549    & 49\,211 \\
        Sent. Length & $21.8\pm 12.8$ & $18.8\pm 12.3$ & $26.5\pm 11.8$ & $23.9\pm 11.3$ \\
        Ent. Types      & 7             & 7             & 5          & 114 \\
        Total Entities  & 27\,751       & 30\,956       & 57\,063    & 279\,796 \\
        Avg. Depth      & 1.34          & 1.29          & 1.10       & 1.88 \\
        Med. Depth      & 1             & 1             & 1          & 2 \\
        Max. Depth      & 6             & 6             & 4          & 6 \\
        \midrule
        \multicolumn{5}{c}{\textbf{Percentage of Entities at Each Nesting Depth}} \\
        \midrule
        0    & 71.77       & 75.81       & 90.29      & 36.45 \\
        1    & 22.98       & 19.98       & 9.45       & 42.73 \\
        2   & 4.61        & 3.73        & 0.25       & 17.83 \\
        3    & 0.55        & 0.44        & 0.00       & 2.83 \\
        4   & 0.08        & 0.02        & 0.00       & 0.15 \\
        5    & 0.01        & 0.01        & 0.00       & 0.00 \\
        \midrule
        \multicolumn{5}{c}{\textbf{Percentage of Entities with Different Span Lengths}} \\
        \midrule
        1 Token & 42.46 & 46.68 & 52.50 & 61.77 \\
        2--4 Tokens & 39.49 & 37.25 & 42.42 & 35.06 \\
        5--9 Tokens & 12.51 & 10.94 & 4.74 & 3.10 \\
        $\geq10$ Tokens & 5.54 & 5.13 & 0.35 & 0.06 \\
        \bottomrule
    \end{tabular}
    \caption{Dataset details, including the percentage of entities at each nesting depth and their distribution by length. Nesting depth refers to the number of hierarchical levels an entity is embedded within another, with 0 indicating a flat entity.}
    \label{tab:datasets}
\end{table*}

\section{Data coverage}
The encodings can represent all nested entity structures in the datasets without error. The only unrecoverable cases with these encodings arise from annotation inconsistencies---such as crossing spans in ACE2004 or inverted boundaries in NNE---rather than from limitations of the encodings themselves. 


\section{Post-processing}
\label{app:post-processing}

Rather than imposing strict constraints during prediction, parsing as sequence labeling ensures well-formedness by applying simple heuristics to ill-formed label sequences. In our setup, we apply the heuristics provided by CoDeLin~\citep{RocVilGomXoveTIC2022}, i.e., (1) in depth-based encodings, if multiple labels are assigned to the same non-terminal node, we retain only the first; (2) if no label is assigned to a node, it is removed from the tree; (3) indices out of range (e.g., -7 in the relative encoding when there are fewer than 7 levels to go up) are changed to the nearest legal index; and (4) in tetra-tagging, if the label sequence specifies an invalid transition, the transition is skipped. Words that remain unattached due to skipped transitions are attached to the lowest non-terminal in the rightmost tree spine.

\section{Model Size and Budget}
We fine-tune three models: BERT Large (340M parameters), BioBERT Large (340M), and RoBERTa Large (355M). Each model is trained with four encoding strategies, across four datasets, and using five random seeds, resulting in a total of 240 training runs. Training was performed on a GPU cluster using NVIDIA A100 GPUs (40GB), with each run executed on a single GPU. The cumulative training time is estimated at approximately 300–360 GPU hours.

\end{document}